\documentclass[runningheads]{llncs}

\usepackage{float}
\titlerunning{AIIB-23 Task2 Solution}
\usepackage{caption}
\usepackage[T1]{fontenc}
\usepackage{orcidlink}
\usepackage{amsfonts}
\usepackage{graphicx}
\begin{document}

\title{Predicting Patient Survival with Airway Biomarkers using nn-Unet/Radiomics}
%
%

\author{
Zacharia Mesbah$^*$\inst{1,4, 5} \and
Dhruv Jain$^*$\inst{1}\orcidlink{0000-0001-7483-409X
} \and
Tsiry Mayet$^*$\inst{1}\orcidlink{0000-0003-3891-1419}\and
Romain~Modzelewski\inst{3}\and
Romain Herault\inst{2}\and
Simon Bernard\inst{1}\and
Sebastien Thureau\inst{4} \\ \and
Clément~Chatelain\inst{1} 
}

\authorrunning{ }
%
\institute{Normandie Univ., INSA Rouen, LITIS, 76801, Saint Etienne du Rouvray, France\\
\and 
Normandie Univ., UNICAEN, ENSICAEN, CNRS, GREYC, 14000 Caen, France \\\and 
Nuclear Medicine Department, Henri Becquerel Cancer Center, Rouen, France\\ 
\and 
Radiotherapy Department, Henri Becquerel Cancer Center, Rouen, France \\ \and
Siemens Healthineers\\
\email{zacharia.mesbah@insa-rouen.fr}\\ 
\email{dhruv.jain@insa-rouen.fr}\\ 
\email{tsiry.mayet@insa-rouen.fr}}

\maketitle   
\def\thefootnote{*}\footnotetext{These authors contributed equally to this work}\def\thefootnote{\arabic{footnote}}
\begin{abstract}
The primary objective of the AIIB 2023 competition is to evaluate the predictive significance of airway-related imaging biomarkers in determining the survival outcomes of patients with lung fibrosis.
This study introduces a comprehensive three-stage approach. 
Initially, a segmentation network, namely nn-Unet, is employed to delineate the airway's structural boundaries.
Subsequently, key features are extracted from the radiomic images centered around the trachea and an enclosing bounding box around the airway. 
This step is motivated by the potential presence of critical survival-related insights within the tracheal region as well as pertinent information encoded in the structure and dimensions of the airway. 
Lastly, radiomic features obtained from the segmented areas are integrated into an SVM classifier. 
We could obtain an \textit{Overall-Score} of 0.8601 for the segmentation in Task~1 while 0.7346 for the classification in Task~2.

\keywords{Survival Analysis  \and Segmentation \and Deep Learning}
\end{abstract}

\section{Introduction}
Quantitative imaging biomarkers (QIB) offer an appealing substitute for measuring lung physiology, including parameters such as the Forced Vital Capacity (FVC)\cite{Raghu2022-gv}.
These subtle changes in airway conditions are highly influential markers for assessing the progression of Idiopathic Pulmonary Fibrosis (IPF). Nonetheless, directly measuring FVC can be susceptible to variations among different laboratories, thereby amplifying the challenges associated with conducting cross-border clinical studies \cite{Flaherty2022-jr}.
The significance of QIB has grown, becoming crucial for diagnosing and characterizing the disease and predicting its future course. Unfortunately, a standardized QIB for diagnosing pulmonary fibrosis is currently absent. 

\subsection{Competition Tasks}
Airway-Informed Quantitative CT Imaging Biomarker for Fibrotic Lung Disease (AIIB) competition aims to  assess the airway bio-markers to enhance the ability to forecast the survival of individuals afflicted by lung fibrosis following a 63-week time frame. The competition has been divided into multiple tasks:
\begin{itemize}
    \item Task 1, Accomplishing precise automated 3D segmentation of lung airways, feasible for utilization as a QIB.
    \item Task 2, Classifying the survival of patients based on their corresponding CT images and associated biomarkers.
\end{itemize}

\subsubsection{Dataset}

A total of 237 samples gathered from multiple centers have been employed for the development and assessment of the module in the presented challenge \cite{Li2022,Nan2022}.
The organizers supplied participants with Training data consisting of CT images paired with biomarker masks ($X = \{a_i, b_i \in  \mathbb{R}^{768 \times 768},\mathbb{R}^{ 512 \times 512} \}$), while the validation data comprised of CT images ($Y = \{a_j \in  \mathbb{R}^{512 \times 512} \}$).
However, the Test set has not been disclosed to participants. 
In Task 2, labels were assigned to 95 patients out of the 120 in the Training Set.
The value of label 0 denotes patients who experienced mortality within 63 weeks, while the value 1 indicates patients who survived. Table~\ref{table:distribution} illustrates the distribution of the dataset for the competition.

\begin{table}[h]
\normalsize
\resizebox{\textwidth}{!}{
\begin{tabular}{|c|c|c|c|}
\hline
                      & \textbf{Train}                                                  & \textbf{Valid}   & \textbf{Test}                                                            \\ \hline
\textbf{No. of Scans} & 120 fibrosis                                                    & 52 fibrosis      & 90 fibrosis + 50 COVID-19 \\ \hline

\textbf{Sizes} & \multicolumn{3}{c|}{768x768, 512x512} \\ \hline
\textbf{Labels}  &  0:36, 1:59 & - &  -         \\ \hline
\end{tabular}}
\caption{Dataset Distribution for the AIIB Challenge}
\label{table:distribution}
\end{table}

\subsubsection{Evaluation}
\begin{itemize}
\setlength\itemsep{1em}
    \item \underline{Task 1}: The airway segmentation task  is evaluated by a combined metric as shown below:\\
    $\mbox{\textit{Overall-Score}} =( loU + Precision+ DBR+ DLR)^* 0.25^* 0.7+(1- leakages)^{\star} 0.3$\\
   where,  \begin{itemize}
   \setlength\itemsep{0em}
       \item IoU: Intersection over union score
\item DLR: Detected length ratio. The length of correctly detected branches in terms of the total length of expert annotations (calculated based on airway centrelines)
\item DBR: Detected branch ratio. The ratio of correctly detected branches in terms of the total branches.
\item Precision: Voxel-wise precision.
\item Leakage: The proportion of total false positive volumes with respect to the ground truth volumes.
   \end{itemize}

   \item \underline{Task 2}: To assess the models created by the participants, a range of classification evaluation metrics is utilized. These metrics comprise of Accuracy, AUC, Sensitivity, Specificity and F1-Score.

\end{itemize}
 
\subsection{No-New-Unet (nn-Unet)}
Currently, nn-Unet~\cite{Isensee2021-bd} stands as the cutting-edge out-of-the-box solution for medical image segmentation, essentially offering optimal Unet configurations determined through the validation scores. Several network configurations are available for assessment, including 2D, 3D Full Resolution, and 3D Low to High Resolution with a cascading network. The remarkable aspect of nn-Unet is its ability to operate seamlessly across diverse image sizes and spacing distributions without requiring resampling or adjustments by the user. It maintains consistency by producing predictions of the same image size and spacing as the input. It examines the supplied training instances and sets up a corresponding segmentation pipeline based on Unet. As a result, the frequent application of nn-Unet in medical imaging competitions is attributed to its exceptional performance and straightforward implementation.

\subsection{Radiomics features} 
Radiomic features are quantitative measurements derived from medical images, acquired primarily through radiological modalities like computed tomography (CT), magnetic resonance imaging (MRI), or positron emission tomography (PET) \cite{Radiomics}.
These features encapsulate diverse attributes of image texture, shape, intensity, and spatial interrelations within designated regions of interest. The goal of radiomic analysis is to convert intricate medical images into an extensive collection of numerical characteristics, which can serve multiple purposes such as disease diagnosis, treatment strategizing, and forecasting outcomes.

\section{Method}
\begin{figure}[h]
    \begin{center}
        \includegraphics[width=1\textwidth]{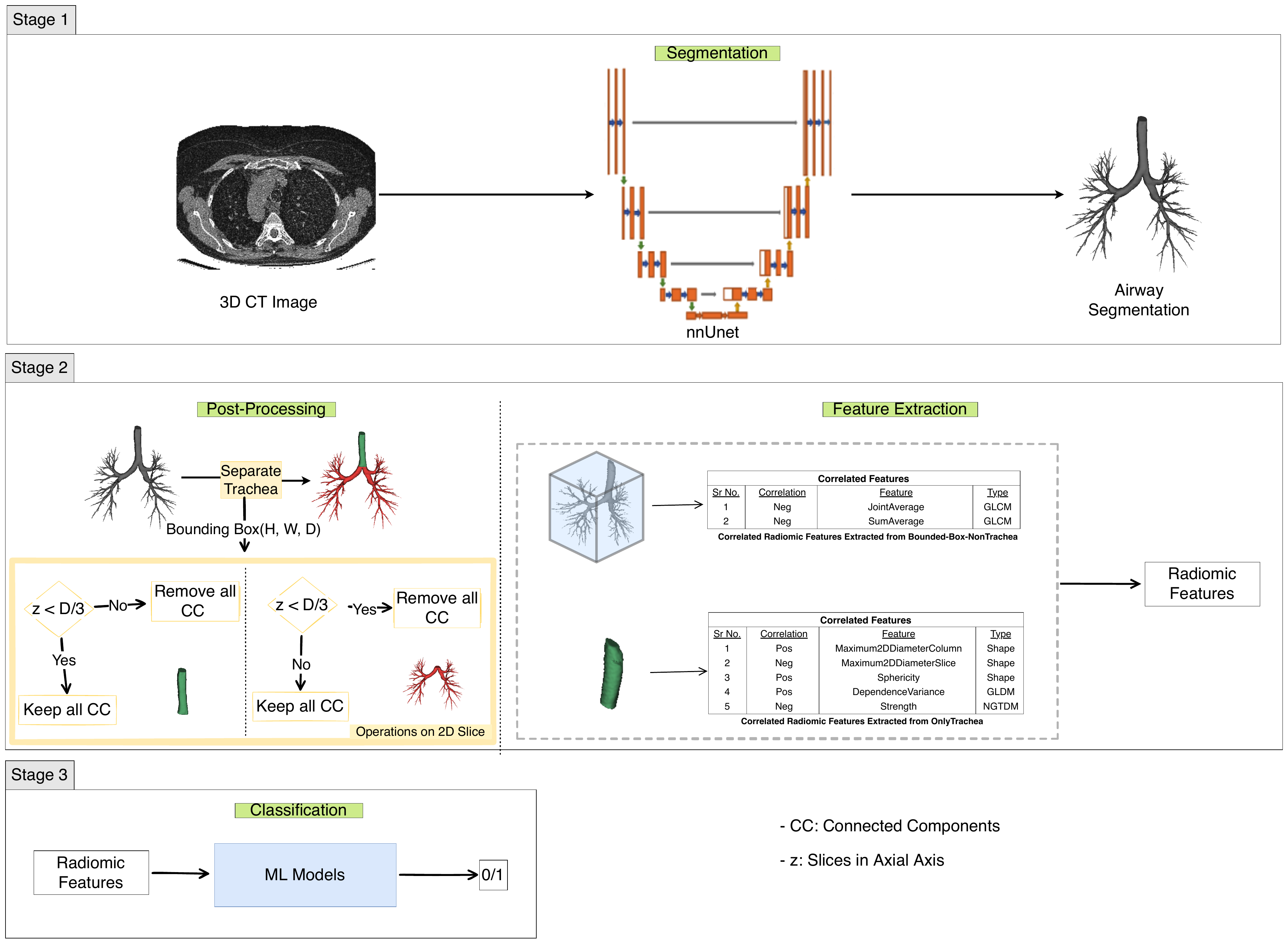}
    \caption{Proposed Approach: Three Stage Pipeline for Survival Classification  }
    \label{fig:proposed-approach}
    \end{center}
    
\end{figure}

We introduce a three-stage methodology in which we first conduct segmentation to obtain the prediction mask for the validation set where they are not available.
Subsequently, in the second stage, we refine the masks through post-processing and extract radiomic features from the scans around the trachea region and the entire airway mask respectively.
These extracted features are then utilized to train an SVM classifier on the training data, followed by prediction on the validation set. Figure~\ref{fig:proposed-approach} shows the pipeline for the 3-Stage approach for the survival prediction. 

\subsection{Stage-1: Segmentation of the Airway Biomarker}
\begin{figure}[h]
    \centering
    \includegraphics[width=\linewidth]{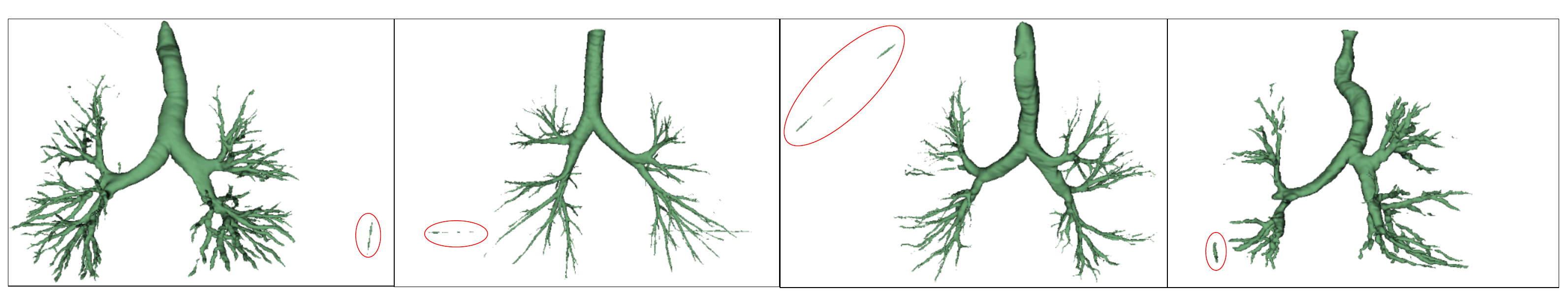}
    \caption{Artifacts shown in the predicted masks before post-processing}
    \label{fig:enter-label}
\end{figure}

During this phase, a direct nn-Unet~\cite{Isensee2021-bd} configuration, specifically 3D\_fullres (an nn-Unet configuration operating on a high image resolution), is trained for 1000 epochs on a single fold. The training process was facilitated using a Tesla~V100~GPU. In the validation phase, we attained an \textit{Overall Score} of 0.8601 for segmentation in Task~1. After evaluating the segmentations, it became evident that most of the samples showcased distinct delineations. Nonetheless, a subset of samples exhibited slight artifacts, which can be easily rectified using post-processing methods. Some of the artifacts are displayed in Figure~\ref{fig:enter-label}. 
Initially, to address minor artifacts, a sequence of morphological closing involving dilation and erosion is implemented on the predicted segmentation mask. Next, we identify the connected components within the segmentation mask, designating the one with the largest area as the main component of airway segmentation.
To ensure the accuracy of the segmentation, we assess the distances between the centroids of each component and the centroid of the primary component. Components whose centroids exceed a predefined threshold are subsequently excluded from the segmentation mask.
Furthermore, it's noteworthy that a section of the fibrous lung is also encompassed within the segmentation. 

\subsection{Stage-2}
\subsubsection{Modified Masks}
Our hypothesis posits that the trachea contains valuable information regarding the patient's survivability, evident in its compressed or elongated appearance. Consequently, there arose a necessity to separate the trachea from other structures of the airway. This led us to create two distinct masks: one for the trachea and another for the remaining airway components. We employ a distinctive approach to isolate the trachea from the airway segmentation. This involves retaining solely the biggest connected component within the upper third of the axial plane and omitting them within the lower third. This method yields the trachea mask, while the process is reversed to obtain the mask for the remaining structure(non-trachea) though we do not use this mask in our proposed approach. 

\subsubsection{Radiomic Features Extration}

Radiomic features calculated across the entire CT volume do not exhibit sufficient relevance to enable effective classification.
To obtain more informative features, we make the assumption that parts of the image closer to the trachea are more relevant to the survival rate.
Therefore, we decide to extract features based on the different mask regions. We used the PyRadiomic~\cite{Radiomics} python library that extracts set of features mentioned below.
\begin{itemize}
    \item First Order Statistics
    \item Gray Level Co-occurrence Matrix - GLCM
    \item Gray Level Size Zone Matrix - GLSZM
    \item Gray Level Run Length Matrix - GLRLM
    \item Gray Level Dependence Matrix - GLDM
    \item Neighbouring Gray Tone Difference Matrix - NGTDM
    \item Shape Based (3D)
\end{itemize}
We extract these features from two masks specifically the trachea mask which is used as is, whereas for the entire airway mask, we enclose it with a minimum bounding box to quantify the characteristics of the overall lung width. Subsequently, we apply Pearson correlation and set a threshold of 0.20 for the trachea, and 0.41 for the airway radiomics.
The resulting performance on the validation set is shown in Table~\ref{table:results}).
Our observations suggest that incorporating these two features improves the model's predictive performance and strengthens its correlation with survival predictions.

\subsection{Stage-3: Classification}
\label{sec:stage-3}
We conduct empirical studies using Support Vector Machine (SVM). Our exploration also encompasses diverse combinations of radiomic features.
We obtained the best model using SVM with RBF kernel, whose parameters(C:8000, gamma:0.01) were evaluated by Grid Search.
We normalized each feature between 0 and 1 independently using min-max normalization computed on the full training data.
We performed 5-fold cross-validation on the training set and used the best-performing model for each of the radiomic features.

\section{Results}
The results show a comparison of the different radiomic features used to predict the classification in the validation set.
Each task has been submitted in the validation phase of the competition to generate the score since the labels were unavailable.
They have been evaluated on the best-found configuration of the support vector machine as mentioned in Section~\ref{sec:stage-3}. Table~\ref{table:results} shows the result on the validation set. Here, the notation Combined(A, B) refers to combination of features extracted from A and from the bounding box of B. 

\begin{table}[H]
\begin{center}
    
\resizebox{\textwidth}{!}{
\begin{tabular}{|c|c|l|l|l|}
\hline
\textbf{Radiomics}                                         & \textbf{Accuracy} & \textbf{F1-Score} & \textbf{Sensitivity} & \textbf{Specificity} \\ \hline
Only Trachea                                               &  0.6923                 &    0.6923               &   0.6923                 &  0.6923               \\ \hline
No Trachea                                                 &   0.5385                & 0.6571                  &  0.8846                  &  0.1923               \\ \hline
 Full Airway &     0.5577              &    0.6849               &  0.9615                  &  0.1538            \\ \hline
\textbf{Combined(Trachea, Airway)}                                         &  0.7308                 & 0.7500                 & 0.8077                  & 0.6538                \\ \hline

Combined(Trachea, non-Trachea)                                        &  0.6731 & 0.6909                 & 0.7308                  & 0.6154                \\ \hline

\end{tabular}}
\caption{Results of SVM-classifier on different Radiomics features in validation set}
\label{table:results}
\end{center}
\vspace{-10mm}
\end{table}
\noindent Based on the findings, we observe that Combined(Trachea, Airway) radiomic features provide the most favorable classification metrics. 
Furthermore, it highlights the significance of the trachea in acquiring the optimal features along with the shape and size of lung features from the bounding box. Although the investigation of various combinations of radiomic features remains open for exploration, it has the potential to yield promising outcomes. An alternative strategy which could be explored involves delving into the isolation of the region below the trachea, forming radiomics features using various smaller airway regions. Additionally, utilizing harmonization among features originating from diverse sites could present an intriguing avenue for attaining greater feature consistency.

\section{Conclusion}
This paper introduces a method for conducting survival analysis on patients with fibrotic lungs, leveraging airway radiomic features. We successfully extracted two significant sets of radiomic features from the trachea and the minimum bounding box of airway, providing valuable insights into patient survival. Our advanced segmentation network effectively segments the airway, and we offer an innovative engineering solution to isolate the trachea mask from the airway mask. It is noteworthy that our proposed technique demonstrated exceptional performance on the leaderboard, positioning it among the top-ranking entries reinforcing its efficacy in the classification task. A remarkable insight drawn from this solution is the identification of significant features within the trachea and the airway structure that warrant further exploration, indicating their potential relevance to patient survival in future studies.

\section{Acknowledgement}
  We would like to thank CRIANN \footnote{Centre des Ressources Informatiques et Applications Numérique de Normandie, France} for their computational resources for this project. The project was also supported by the MINMACS Région Normandie excellence label and  "ANR-20-LCV1-0009"-ANR-LabCom L-Lisa. We would also like to express our gratitude to Fabian Isensee for his assistance with the Docker submission using nnunetv2.

%
%
%
\bibliographystyle{unsrt}
\bibliography{mybibliography}

\end{document}